\documentclass[twoside]{article}

\usepackage{hyperref}
\usepackage[preprint]{aistats2027}
\usepackage{amsfonts}
\usepackage{amsmath}
\usepackage{amsthm}
\usepackage{xcolor}
\usepackage{graphicx}
\usepackage{booktabs}
\usepackage{natbib}

\newtheorem{theorem}{Theorem}
\newtheorem{corollary}{Corollary}

\newcommand{\x}{\mathbf{x}}
\newcommand{\bv}{\mathbf{v}}
\newcommand{\n}{\mathbf{n}}
\newcommand{\m}{\mathbf{m}}
\newcommand{\X}{\mathbf{X}}
\newcommand{\I}{\mathbf{I}}
\newcommand{\La}{\mathbf{L}}
\newcommand{\Q}{\mathbf{Q}}
\newcommand{\y}{\mathbf{y}}
\newcommand{\z}{\mathbf{z}}
\newcommand{\sol}{\mathbf{c}}
\newcommand{\be}{\boldsymbol{\beta}}
\newcommand{\bt}{\boldsymbol{\theta}}
\newcommand{\bg}{\boldsymbol{\gamma}}
\newcommand{\bep}{\boldsymbol{\epsilon}}
\newcommand{\e}{\mathbf{e}}

\newcommand{\Sw}{\mathbf{S}}

\newcommand{\argmax}[1]{\underset{#1}{\textrm{argmax}}\,\,}
\newcommand{\sgn}{\textrm{sgn}}
\newcommand{\clip}{\textrm{clip}}
\newcommand{\cone}{\textrm{cone}}
\newcommand{\conv}{\textrm{conv}}
\newcommand{\cut}{\textrm{cut}}
\newcommand{\Pb}{\bar{P}}

\begin{document}



\twocolumn[

\aistatstitle{Acquisition for Spherical-Linear Bayesian Optimization on the Cube}

\aistatsauthor{Nathan Wycoff}

\aistatsaddress{
Department of Mathematics and Statistics \\
University of Massachusetts Amherst} 
]

\begin{abstract}
Recent work demonstrated that executing Bayesian Optimization (BO) using simply a linear surrogate with a clever transformation is enough to approximate the performance of the standard Gaussian-Process (GP) based version on high-dimensional test functions.
When the decision region is a sphere, this strategy admits rapid exact acquisition.
However, it is the unit cube we encounter most often in the BO literature, and in this setting a standard approach, namely a multi-start quasi-Newton method, is employed for solving the acquisition subproblem.
In this article, we will discuss a series of results, both positive and negative, regarding the complexity of global resolution of various acquisition subproblems over the unit cube.
On the basis of this discussion, we will propose a new acquisition function which is globally optimizable using significantly less execution time than the incumbent approach while maintaining or improving performance. 
The resulting method is competitive with GP-based BO in terms of accuracy per evaluation, and is superior in terms of accuracy per walltime on black-boxes of low to moderate computational intensity.
\end{abstract}

\section{Introduction}

Prior to 2026, the idea of using a global linear model for BO had never been seriously considered because a nontrivial linear map increases monotonically in some direction, such that if the true function is indeed a line, its optimum must lie on a boundary. 
The BO machinery, understanding this, proposes points on the boundary as well, and cannot search the entire interior of the space.
What's worse, the bounds solicited from the practitioner are commonly loose, such that they are somewhat beyond where the reasonable values of a parameter are likely to be to ensure that the optimum is in the search region. 

On the other hand, boundary-seeking behavior has classically been reported \citep{binois2022survey} also for Gaussian Process surrogates\footnote{Though the prevalence of this has recently been disputed, see \citet{hvarfner2024vanilla}.}, prompting many to consider methods by which it might be mitigated.
The most straightforward resolution might be just  reducing the dimension either via variable selection \citep[e.g.,][]{shen2023computationally}, linear dimension reduction \citep{wang2016bayesian} or some hybrid \citep{eriksson2021high}.
More sophisticated strategies include a cylindrical transformation as deployed by BOCK \citep{oh2018bock},  or restricting the search to some subset of the space as deployed by TuRBO \citep{eriksson2019scalable}, which attempt to downplay the boundary of the space while remaining in high dimension.
Set against this history, the project of \citet{doumont2026still} is a radical departure: they go in the exact opposite direction, and propose a transformation which puts \textit{all} points on the boundary.
%Hence the intuitive conceptual appeal of choosing points on a sphere:
In particular, they propose to use the inverse transformation of \textit{stereographic projection} from cartography which maps the sphere to the plane.
Perhaps at first counterintuitive, we can capture the essence of this approach by misquoting \textit{The Incredibles}: ``When everyone is on the boundary, no one is''.
But while the \textit{sphere} doesn't have a boundary, the preimage of the unit cube under stereographic projection is not the whole sphere, and it \textit{does} have edges, which as we will see cause their own issues.
%We will show in this article that in particular it is (almost) given by the intersection of the whole sphere with a polyhedral cone.
Further, since the sphere and unit cube are not homeomorphic, there is no alternative continuous transform that achieves complete coverage of the sphere which might extirpate this issue.

\begin{figure*}
    \centering
    \includegraphics[width=0.95\linewidth]{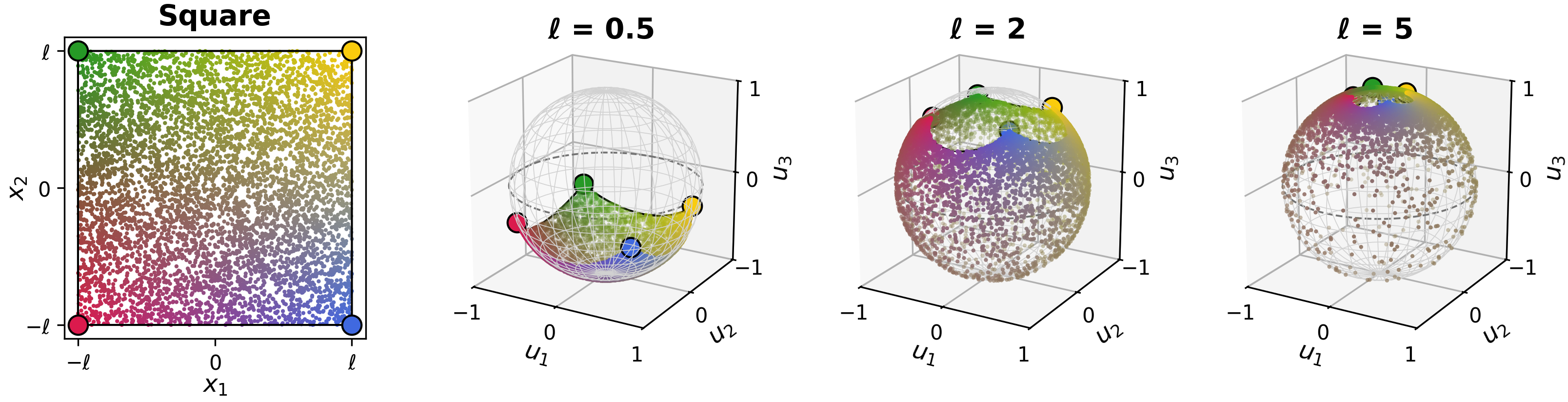}
    \caption{\textbf{Search Region for BO in Spherical Coordinates.}
    \textit{Left Panel:} Random points in the unit cube $[-\ell,\ell]^2$, with coloration shared across panels.
    \textit{Remaining Panels:} The image of $[-\ell,\ell]^2$ under $P$ for various settings of $\ell$, showing the shape of the feasible region. 
    As $\ell$ grows, the preimage grows, the space around the origin is stretched, and the corners of the unit cube are compressed to the north pole.
    }
    \label{fig:stereo_row}
\end{figure*}

In this article, we will study the prospects for efficiently solving the acquisition subproblem when performing Spherical-Linear Bayesian Optimization (SLBO).
In particular, our contributions are:
\begin{enumerate}
    \item We characterize the preimage of the unit cube under stereographic projection, finding it to be in particular (almost) given by the intersection of the whole sphere with a polyhedral cone.
    \item We develop an efficient global algorithm for performing Thompson sampling (TS) for SLBO.
    \item We show that SLBO-TS underexploits in high dimension, and develop a new acquisition rule which we term \textit{Surface Thompson Sampling}.
    \item We demonstrate that global resolution of the Expected Improvement and Upper Confidence Bound acquisitions is in general intractable for SLBO.
    \item We show the advantages of our new acquisition strategy on a battery of numerical experiments.
\end{enumerate}

\subsection{Notation}
If $\mathcal{A}\subseteq\mathbb{R}^D$ is a set and $a\in\mathbb{R}$, $\mathbf{b}\in\mathbb{R}^D$, then $a\mathcal{A}+\mathbf{b} := \{a\x+\mathbf{b} : \x\in\mathcal{A}\}$.
If $f: \mathcal{D}\to\mathcal{C}$ is a function, we will denote by $f^{-1}(\mathcal{A})$ the preimage of $\mathcal{A}\subseteq\mathcal{C}$.
We denote the sign function $\sgn(x)\in\{-1,0,1\}$, which takes value $0$ if $x=0$, and 
%its Fenchel dual % oh no wait, clip is dual to softthresh.
the clip function $\clip(x,a,b)$, which just returns $x$ if $x\in[a,b]$ and if not returns the closer endpoint.
Given some subset $\mathcal{A}$ of a linear space, its \textit{convex hull} $\conv(\mathcal{A})$ gives the set of all finite convex combinations of elements of $\mathcal{A}$.
In general, given a linear space, a \textit{convex cone} $\mathcal{C}$ is a subset of that space closed under nonnegative linear combination.
Given some vectors $\mathbf{a}_1,\ldots,\mathbf{a}_K$, their \textit{polyhedral cone}, denoted $\cone(\mathbf{a}_1,\ldots,\mathbf{a}_K)$, denotes the cone composed of nonnegative linear combinations of those vectors; we use similar notation $\cone(\mathbf{A})$ to denote the polyhedral cone induced by $\mathbf{A}$'s columns.
%Cones generated in this manner are called \textit{polyhedral cones}.
In general, a cone $\mathcal{C}$ in a linear space induces a partial ordering on that space, where $\mathbf{a}\leq_{\mathcal{C}}\mathbf{b}$ exactly when $\mathbf{b}-\mathbf{a}\in\mathcal{C}$.
$S_D$ denotes the sphere of dimension $D$, considered in this article extrinsically as a subset of $\mathbb{R}^{D+1}$.

\section{Background}

We begin with a general overview of Bayesian Optimization before describing the recent line of work focusing on its behavior in high dimension.

\subsection{Bayesian Optimization}

Bayesian Optimization \citep{garnett_bayesoptbook_2023} involves finding the minimum of a function $f:\mathcal{X}\to\mathbb{R}$ by alternating between 1) fitting a statistical model, known in this context as a \textit{surrogate} \citep{gramacy2020surrogates}, to observed input-output pairs, 2) querying that model for a new point $\x^*$ at which to evaluate $f$, then 3) evaluating the objective and adding $(\x^*, f(\x^*))$ to the observed data.
This article is primarily concerned with step (2), which we call the \textit{acquisition subproblem}.
Typically, it itself is constituted by the resolution of an auxiliary optimization problem $\x^* = \argmax{\x\in\mathcal{X}} \alpha(\x)$.
This function $\alpha$ whose maximum defines our next design point is called an \textit{acquisition function}.
The most popular acquisition functions are Expected Improvement \citep[EI;][]{jones1998efficient}, the Upper Confidence Bound \citep[UCB;][]{srinivas2009gaussian}, and a posterior function draw which induces Thompson sampling \citep{thompson1933likelihood}.
In the typical GP-based BO, these acquisition functions are highly multimodal, and it is not possible to reliably determine their global optimum for $\mathcal{X}$ of even moderate dimension. 
In practice, the typical strategy is to conduct multiple quasi-Newton searches, usually using L-BFGS-B, initialized with a diversity of starting values.

\subsection{Spherical-Linear Bayesian Optimization}

Recently, a number of articles have challenged the classical understanding of BO in high dimension.
\citet{hvarfner2024vanilla} and \citet{xu2025standard} showed that when properly implemented, ``vanilla'' BO with GPs can perform better than was sometimes observed. 
As we described in the introduction, \citet{doumont2026still} propose instead to pass design points $\x\in[-\ell,\ell]^D$ through the transformation:
\begin{equation*}
    P: \mathbb{R}^D \to S_D\setminus\{(\mathbf{0},1)\};
    \quad
    \x \mapsto \frac1{\Vert\x\Vert^2+1}
    \begin{bmatrix}
        2\x^\top & \Vert\x\Vert^2-1
    \end{bmatrix}^\top
    \,.
\end{equation*}
It is easily verified that the stereographic projection which is its inverse $P^{-1}$ is given by:
\begin{equation*}
    P^{-1}: S_D\setminus\{(\mathbf{0},1)\}\to\mathbb{R}^D; (\y,s)\mapsto \frac{\y}{1-s} \,,
\end{equation*}
where $(\y,s)$ are extrinsic coordinates in $\mathbb{R}^{D+1}$.
Subsequently, SLBO assumes that there is some coefficient vector $\be\in\mathbb{R}^{D+2}$ such that $y_i|\be,\x_i\sim N(\Pb(\x_i)^\top\be, \sigma^2)$, where $\Pb(\x):=[P(\x)^\top, 1]^\top$.
After $N$ observations organized into $\X\in\mathbb{R}^{N\times D}$ and $\y\in\mathbb{R}^N$, the posterior on $\be$ is given by the standard Bayesian linear analysis:
\begin{align*}
    & \be|\y,\X\sim N\left(
    \m, \Sw\right),
    \qquad\textrm{with}
    \qquad \be\sim N(\mathbf{0}, \Lambda);
    \\ & 
    \Sw =
    \left(
    \frac{1}{\sigma^2}\Pb(\X)^\top \Pb(\X)
    + \Lambda^{-1}
    \right)^{-1}
    \hspace{-0.5em};\,
    \mathbf{m} = \frac{1}{\sigma^2}\Sw \Pb(\X)^\top\y\,.
\end{align*}

In the code accompanying the original article, \citet{doumont2026still} simply used the standard multi-start quasi-Newton strategy to approximately optimize expected improvement.
However, in a very recent work, \citet{fan2026search} show that when $\mathcal{X} = S_D$, UCB, EI and Thompson sampling are efficiently computable.
They operationalize this observation to perform BO in the latent space of a neural model to great effect.
Unfortunately, in many other practical situations, $\mathcal{X}$ is not naturally expressible as a sphere. 
Most commonly, $f$ represents some system with $D$ independent levers to manipulate, which mathematically is representable as a box isomorphic to $[-1,1]^D$.
Their results do not apply to this setting, which is rather the subject of study in this article.

\section{The Preimage of the Unit Cube under Stereographic Projection}

%We are used to working with the $[0,1]^D$ box in BO, but it will be important to use one centered at $0$, perhaps $[-\ell,\ell]^D$, instead, where $\ell$ is a global length-scale parameter.
Figure \ref{fig:stereo_row} shows the image of $[-\ell,\ell]^2$ under $P$ in dimension 2 for various $\ell$. 
We notice that the portion of the sphere covered by $P([-\ell,\ell]^D)$ grows with $\ell$, which has the dual effect of pushing all the boundaries together and exaggerating the space near the origin.
In this section, we will characterize this region.
%as a subset of $\mathbb{R}^{P+1}$.

%We begin by giving an algebraic formulation for the preimage of $[-\ell,\ell]^D$ under stereographic projection.

%Though this expression is exact it does not give us much intuition. 
%To this end, we will develop a union bound on it%, and start by considering the behavior of a single coordinate $z_i$.
%For each coordinate $z_i$, we have that $P(|z_i|>\ell) = P(|\tau|>\ell\sqrt{D})$ where $\tau$ is a $t$-distributed random variable with $D$ degrees of freedom.
%The union bound therefore gives us
%\begin{equation}
%    1-P(\z\in[-\ell,\ell]^D) \leq 2D P(t>\ell\sqrt{D}).
%\end{equation}
%The $t$-cdf is given in terms of special functions, so we now approximate it.

%Take some coordinate $z_i$.
%If $|z_i|>\ell$, then 
%\begin{align*}
%    \theta_i > \ell(1-\theta_{D+1})
%    %\iff 
%    %\begin{bmatrix}
%    %    \e_i^\top & \ell
%    %\end{bmatrix}
%    %\bt \geq \ell
%    \iff 
%    \frac{
%    \begin{bmatrix}
%        \e_i^\top & \ell
%    \end{bmatrix}
%    }{\sqrt{1+\ell}}
%    \bt \geq \frac{\ell}{\sqrt{1+\ell}}
%    \,,
%\end{align*}
%and we define the unit vector $\n$ and constant $h$ such that we are interested in computing $P(\n^\top\bt>h)$.
%But by the rotational symmetry of $\bt$'s uniform distribution, this is equal to $P(\e_1^\top\bt>h)$.
%By \citet{vershynin2019high}'s Proposition 3.3.10, $P(u_1)\propto (1-u_1^2)^{-\frac{D-2}{2}}$ such that:
%\begin{equation*}
%    P(|z_i|>\ell) = \int_{\frac{l}{\sqrt{l}+1}}^1(1-u^2)^{-\frac{D-2}{2}} du
%\end{equation*}

\begin{theorem}
    Let $\Omega:= P([-\ell,\ell]^D)$.
    Then for $\epsilon>0$ given in the proof, $\Omega$ is expressible as:
    \begin{align*}
        \{ (\y,s)\in\mathbb{R}^{D+1} : 
        \Vert(\y,s)\Vert_2=1, |y_d| 
        \leq\ell(1-s),s\leq1-\epsilon\} 
        \,.
    \end{align*}
    \label{thm:image}
\end{theorem}
\begin{proof}
Appendix \ref{ap:algebra}.
\end{proof}

These algebraic relationships come in two flavors: the sphericity constraint is a quadratic equality constraint, while all the others are linear constraints.
This suggests the following geometric picture.

\begin{corollary}
    Let $\n = [\mathbf{0},1]$ be the north pole of the unit sphere,
    and 
    \begin{equation}
        \mathcal{H} = 
        \cone(
        \{
        (\ell \bv, -1) : \bv\in\{-1,1\}^D
        \}
        )
        \,.
    \end{equation}
    Then $\Omega$ is expressible as the intersection of that cone shifted by the north pole and the sphere, minus the north pole:
    \begin{align*}
        & \Omega \cup \{\n\}
        = (\mathcal{H}+\n)\cap S_{D} \,.
    \end{align*}
    \label{cor:poly}
\end{corollary}
\begin{proof}
Appendix \ref{ap:geometry}.
\end{proof}

\begin{figure}
    \centering
    \includegraphics[width=0.95\linewidth]{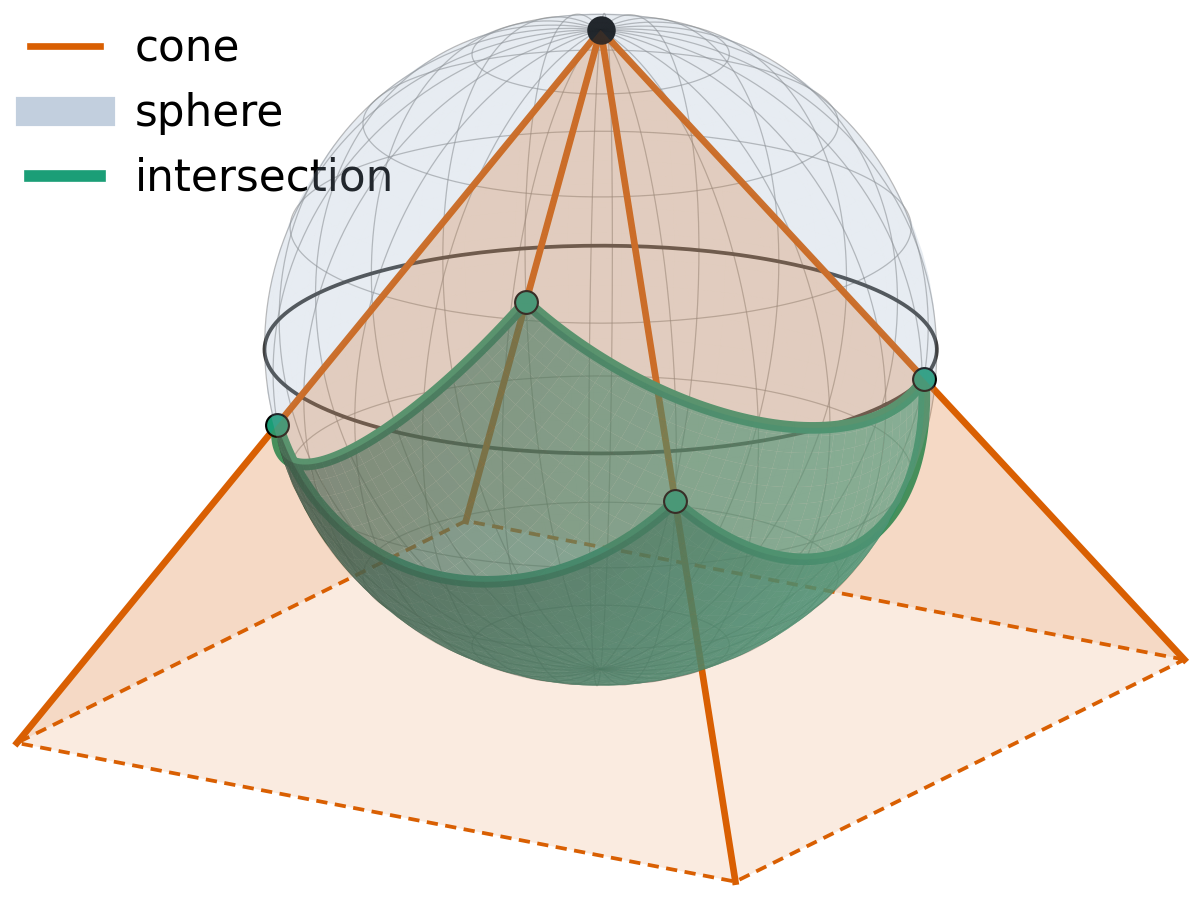}
    \caption{\textbf{Illustration of Corollary \ref{cor:poly}}.
    The image of the constraint set $[-\ell,\ell]^D$ under the inverse stereographic transformation, shaded in green, is given by the intersection of the polyhedral cone $\mathcal{H}$ originating at the north pole (orange) and the sphere $S_D$ (gray), minus the point at the north pole.
    }
    \label{fig:polyhedron}
\end{figure}

This result tells us that we \textit{almost} have $P([-\ell,\ell]^D) = (\mathcal{H}+\n)\cap S_D$, except this intersection gives us one more point than we want: namely, the north pole, since
$0\leq \ell(1-1)=0$, $1\leq 1$ and $\Vert\n\Vert = 1$.
This is why we needed the constraint $s\leq 1-\epsilon$ in Theorem \ref{thm:image}: any positive $\epsilon$ is sufficient to exclude $\n$, and choosing $\epsilon\leq\frac{2}{D\ell^2+1}$ is sufficient for making sure all of $[-\ell,\ell]^D$ is included.
Figure \ref{fig:polyhedron} illustrates $P([-\ell,\ell]^D)$, $\mathcal{H}$, $S_D$, and $\n$ in dimension 2.

From a practical perspective, this means that we can formulate the acquisition problems either as being just box-constrained, but pushed through the nonlinear warping $P$, or as constrained by the quadratic constraint $\Vert(\y,s)\Vert=1$ together with linear constraints.
To illustrate, consider arguably the simplest acquisition function, predictive variance maximization:
\begin{align*}
    \max_{\x\in[-\ell,\ell]^D}
    \sigma^2(\x) =
    \max_{\x\in[-\ell,\ell]^D}
    \Pb(\x)^\top\Sw \Pb(\x)\,,
\end{align*}
where $\Sw$ is the posterior covariance for the linear regression in spherical-space.
Our results in this section mean this may be re-expressed as:
\begin{align*}
    & \qquad\qquad\quad\max_{\x\in\mathbb{R}^{D+1}}
    \begin{bmatrix}\x\\1\end{bmatrix}^\top\Sw\begin{bmatrix}\x\\1\end{bmatrix}
    \\ & 
    s.t. \quad \Vert\x\Vert^2=1; \quad 
    \mathbf{H}\mathbf{x} \leq \ell\mathbf{1}; \quad 
    \mathbf{e}_{D+1}^\top \x \leq 1-\epsilon\,,
\end{align*}
with 
$
    \mathbf{H} = 
    \begin{bmatrix}
        \I & \ell\mathbf{1} \\
        -\I & \ell\mathbf{1}
    \end{bmatrix}
$. 
This is recognized as a Quadratically Constrained Quadratic Program (QCQP).
With this recognition comes both good and bad news.
The good news is that this places variance-maximization for SLBO into a standard mathematical framework, amenable to general-purpose optimization software such as Gurobi.
The bad news is that QCQPs, as a general class, are computationally difficult: the unit-sphere constraint is inherently nonconvex,
and though it in-and-of-itself is tractable as shown by \citet{fan2026search},
the addition of the otherwise innocuous linear constraints is enough to make this problem intractable as we will see later in this article.
We tested the commercial Gurobi solver and the open-source SCIP \citep{scip} and COIN-OR \citep{lougee2003common} QCQP solvers, but none was able to improve on the incumbent multi-start strategy.
%But first, we give some more good news.

\subsection{Proportion of $S_D$ covered by $P([-\ell,\ell]^D)$}
Given that solving the acquisition problem over the whole sphere is simple, a natural question is how big the gap is between $P([-\ell,\ell]^D)$ and $S_D$.
After all, since $P([-\ell,\ell]^D)\subset S_D$, if $\bt^*$ is the optimum of an acquisition function over $S_D$ and happens to lie in $P([-\ell,\ell]^D)$, then $P^{-1}(\bt^*)$ must be the optimum also over $[-\ell,\ell]^D$.
To this end, we study the proportion of points on the sphere whose preimage under $P$ lies in $[-\ell,\ell]^D$.
Assume we draw a uniform coefficient $\bt$ on the sphere. 
Its backtransformation $\z:=P^{-1}(\bt)\in\mathbb{R}^D$ has density given by:
\begin{equation*}
    f(\z) = 
    f_{\bt} (P(\z))
    \left|
    J_{\z} P(\z)
    \right|
    \,.
\end{equation*}
$f_{\bt} (P(\z))$ is constant since we are drawing $\bt$ uniformly.
\citet{bell2024adaptive}'s Equation 2.4 gives us that the Jacobian term is proportional to 
$\left(
1+\Vert\z\Vert^2
\right)^{-D}$.
This is almost the density of a multivariate $t$ distribution; it is in fact $\frac{\bg}{\sqrt{D}}$ if $\bg$ is multivariate $t$ with $D$ degrees of freedom.
Therefore, since $\bg \overset{d}{=} \frac{\bep}{\rho/\sqrt{D}}$ with $\bep$ a $D$-vector of iid standard normals and $\rho$ an independent $\chi_{D}$ variate, $\z \overset{d}{=} \frac{\bep}{\rho}$.
Therefore, conditional on $\rho$, the probability of a given coordinate $d$ lying in $[-\ell,\ell]$ is given by:
\begin{equation*}
    %P(|z_d|\leq\ell) = 
    P(|\epsilon_d|\leq \ell\rho)
    = P(\epsilon_d\leq \ell\rho)-P(\epsilon_d\leq-\ell\rho)
    = 2\Phi(\ell\rho)-1
    \,,
\end{equation*}
%The conditional probability of all coordinates (independent given $\rho$) is thus simply 
and thus 
\begin{equation*}
    P\left(\bep\in\rho[-\ell,\ell]^D|\rho\right)
    = (2\Phi(\ell\rho)-1)^D\,,
\end{equation*}
leaving us to compute $P(\z\in[-\ell,\ell]^D) = \int P\left(\bep\in\rho[-\ell,\ell]^D|\rho\right) P(\rho) d\rho
= 
\mathbb{E}_{\rho} 
[(2\Phi(\ell\rho)-1)^D]
$.

\begin{figure}
    \centering
    \includegraphics[width=0.995\linewidth]{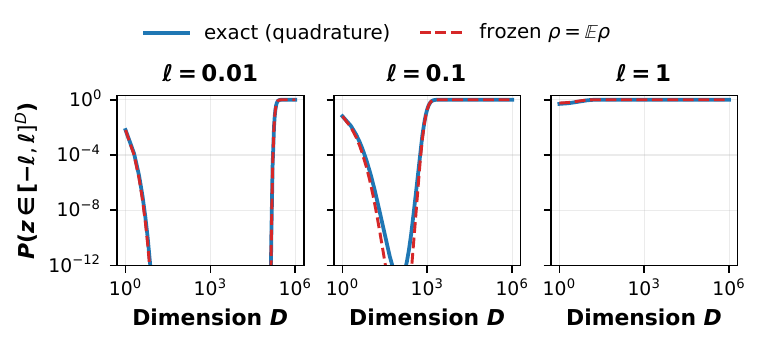}
    \caption{\textbf{Probability a Uniform Point on Sphere is Box-Feasible}.
    The solid blue line gives a numerical estimate of the true relationship derived via Gaussian quadrature, while the dotted red line gives an estimate obtained from a thin shell approximation.
    We see that the relationship is not monotonic in $D$.
    }
    \label{fig:feas}
\end{figure}

We solve this $1$-dimensional integral via Gaussian quadrature and plot the results in Figure \ref{fig:feas}, alongside an approximation given by taking $\rho$ to be equal to its expectation.
Intriguingly, we find that this probability is not monotonic in $D$, and in fact first decreases before later increasing. 
This makes it difficult to know when we can expect the backprojection to work, and motivates the study of exact solutions to this problem.
Nevertheless, it is promising that this probability tends to 1 as $D$ diverges. 
However, in practice optima will not be uniformly distributed, and as we will show computation is still likely superpolynomially difficult in the worst case.

\section{Thompson Sampling is Computationally Easy}

\begin{figure*}
    \centering
    \includegraphics[width=0.995\linewidth]{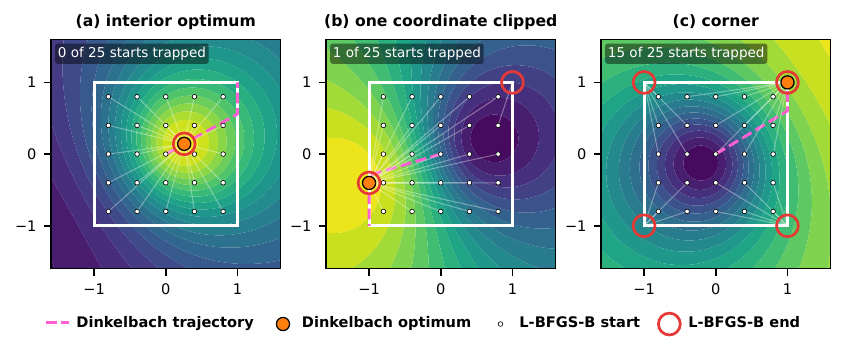}
    \caption{\textbf{Necessity of Dinkelbach Iteration.}
    These panels show three random spherical-linear functions on $[-1,1]^2$, as well as the performance of 25 L-BFGS-B runs (initializations in white linked to solutions in open red circles) and our proposed Dinkelbach iteration (path in pink and solution in closed orange circle).
    Panel (a) shows a function which is concave within the unit cube and which is amenable to a standard quasi-Newton search.
    In Panel (b), the desired maximum is almost in the box, but the minimum is too, and points on the other side cannot cross it. 
    Panel (c) shows the local search methods failing to maximize a convex function unless they are propitiously initialized. 
    In all cases, the Dinkelbach iteration is successful, as anticipated by its global convergence properties.
    }
    \label{fig:ts_cost}
\end{figure*}

Since the intercept $\beta_{D+2}$ does not depend on $\x$, it's easy to see that to effect an unconstrained optimization of the target linear function over the sphere, we choose in sphere-coordinates merely the normalized regression coefficient $\frac{\be_{1:D+1}}{\Vert\be_{1:D+1}\Vert_2}$. 
Pushing that through the stereographic projection yields 
$
\frac{\be_{1:D}}{\Vert\be_{1:D+1}\Vert-\be_{D+1}}
$.
But $\frac{\be_{1:D+1}}{\Vert\be_{1:D+1}\Vert_2}$ may not lie in the preimage of the unit cube, or equivalently, 
$
\frac{\be_{1:D}}{\Vert\be_{1:D+1}\Vert-\be_{D+1}}
$ 
may be outside of the box.
We may be tempted to try to project it onto the box after the fact.
But Figure \ref{fig:stereo_row} shows us that this set has protrusions such that computing the projection of a point on the sphere but outside of the preimage using typical methods would be difficult as it is clearly non-Chebyshev \citep[Chapter 28]{bauschke2020convex}.
%Our intuition would like to say that the preimage is nonconvex, but this is clearly true since even the full sphere itself is nonconvex.
%We will be more precise and say that the preimage is locally non-Chebyshev when viewed as embedded within the sphere, which makes it a seemingly complicated problem to solve.
Luckily, some operations remain tractable in this setting, as we delineate in the remainder of this section.

Say that we have sampled $\be$.
Then generating a Thompson sample means solving $ \argmax{\x\in[-\ell,\ell]^D} \be^\top \Pb(\x)  $
\begin{align*}
   = \argmax{\x\in[-\ell,\ell]^D}
    \frac{2\sum_{d=1}^D \beta_d x_d + (\Vert \x \Vert^2-1)\beta_{D+1}}{\Vert \x\Vert^2 + 1} \,.
\end{align*}
This is recognized as the ratio of two quadratics, which can be resolved via the techniques of \textit{Fractional Programming} \citep{schaible1995fractional}.
This involves solving a sequence of \textit{Dinkelbach subproblems} \citep{dinkelbach1967nonlinear} which are maximization problems subtracting a multiple $q$, to be determined shortly, of the denominator from the numerator, giving in our case:
\begin{align*}
     \x^*(q) \gets \argmax{\x\in[-\ell,\ell]^D}
    (\beta_{D+1}-q) \Vert \x \Vert^2 +2\sum_{d=1}^D \beta_d x_d \,.
\end{align*}
Though only concave when $\beta_{D+1}\le q$, it is nevertheless isotropic and constrained by a rectangular domain and therefore reduces to resolving the coordinate problems independently regardless of convexity:
\begin{align*}
    x^*_{d}(q) \gets \argmax{x\in[-\ell,\ell]}
    (\beta_{D+1}-q)
    x^2
    + 2\beta_d x \,.
\end{align*}
This has stationary point $\frac{\beta_d}{q - \beta_{D+1}}$, and if $\beta_{D+1}<q$, optimization consists of getting as close to it as possible.
Otherwise, we want to move in the same direction as $\beta_d$ until we hit the boundary, yielding:
\begin{align*}
    x^*_d(q) \gets 
    \begin{cases}
    \clip\left(
    \frac{\beta_d}{q-\beta_{D+1}}
    ,-\ell,\ell
    \right)
    & \beta_{D+1}<q
    \\
    \ell\,\sgn(\beta_d) & \beta_{D+1}\geq q
    \,.
    \end{cases}
\end{align*}

Dinkelbach's method then proceeds by iterating:
\begin{enumerate}
    \item $q^t \gets \frac{2\sum_{d=1}^D \beta_d x^t_d + (\Vert \x^t \Vert^2-1)\beta_{D+1}}{\Vert \x^t\Vert^2 + 1}$
    \item $\x^{t+1}\gets \x^*(q^t)$.
\end{enumerate}
By classical results, this iteration converges superlinearly in $t$ to a global optimum.
Figure \ref{fig:ts_cost} illustrates how this iteration can find the global optimum even in settings where local search methods get trapped in local optima.
In our experiments, it never took more than 8 iterations to reach convergence, such that it is both much faster and has higher solution quality than the incumbent approach.

\subsection{A Pathology in Large Dimension}

We have shown that Thompson sampling is computationally tractable, but we have not discussed its usefulness. 
To this end, we consider, with $\be\sim N(\m,\Sw)$:
\begin{align*}
    &
    \argmax{\x\in[-\ell,\ell]^D}
    \be^\top \Pb(\x)
    =
    \argmax{\x\in[-\ell,\ell]^D}
    \left(
    \m + \Sw^{\frac{1}{2}} \bep
    \right)^\top
    \Pb(\x)
    \\ & = 
    \argmax{\x\in[-\ell,\ell]^D}
    \frac1{\Vert\bep\Vert}\m^\top \Pb(\x) + \frac{\bep}{\Vert\bep\Vert}^\top\Sw^{\frac{1}{2}}\Pb(\x)
    \,.
\end{align*}
where $\Sw^{\frac12}$ is any matrix with $\Sw^{\frac12}(\Sw^{\frac12})^\top=\Sw$, such as its Cholesky factor and $\bep$ is a vector of $D+2$ iid standard normals.
Under this assumption, $\Vert\bep\Vert\sim \chi_{D+2}$ \citep[Chapter 3]{graybill1976theory}, and we have:
\begin{align*}
    &
    \mathbb{E}[\Vert\bep\Vert] = \frac{\sqrt{2}\Gamma(\frac12(D+3))}{\Gamma(\frac{D+2}{2})}
    \overset{D\to\infty}{\sim}
    \sqrt{D+\frac32}
    \\ & 
    \mathbb{V}[\Vert\bep\Vert] = 
    D+2 - \mathbb{E}[\Vert\bep\Vert]^2
    \overset{D\to\infty}{\to}
    \frac12\,.
\end{align*}
Therefore, $\Vert \bep \Vert$ asymptotically has its probability mass in a constant-size interval around a growing mean:
\begin{align*}
    \be^\top \Pb(\x)
    \overset{D\textrm{ large}}{\approx}
    \frac{\bep^\top}{\Vert\bep\Vert} (\Sw^{\frac12})^\top
    \Pb(\x)
    :=
    \z^\top \Pb(\x)
    \,,
\end{align*}
where $\z$ is an angular Gaussian \citep{wang2013directional} scaled by $\Sw$.
%\todo{$\Sw^{\frac12}\bep/\Vert\bep\Vert$ is not unit-norm, so it is not exactly an angular Gaussian.}

SLBO was motivated in the context of high dimension, but we see here that TS in this context totally ignores mean information, and simply suggests directions which are not pinned down by the posterior covariance matrix $\Sw$.
In a linear model, the covariance matrix is independent of $\y$, and TS therefore asymptotically disregards the blackbox output.
Early experiments showed that a BO loop built on this acquisition made little progress in optimizing the objective.

\subsection{Thompson Sampling on a Surface}

While it is tempting to simply move on from TS and consider other common acquisition strategies, as we will show below, TS is by far the easiest of these to solve.
We therefore in this section propose a method to repair the issue.
Since the source of the issue is that the norm of the iid vector $\bep$ grows, we propose to simply control this norm using a parameter $\kappa>0$ when selecting the next point $\x^*$:
\begin{align*}
    \tilde{\be} := \m + \kappa \Sw^{\frac12} \frac{\bep}{\Vert\bep\Vert}
    \,; \quad
    \x^* = \argmax{\x\in[-\ell,\ell]^D}
    \tilde{\be}^\top \Pb(\x)
    \,.
\end{align*}
The form of the acquisition subproblem once the coefficients are sampled is identical to that of standard TS, being again a linear optimization, and amenable to the same iteration, meaning it can also be globally optimized with little computational burden.
Under this sampling scheme, we have the following expression for the $\Sw$-Mahalanobis norm of $\tilde{\be}-\m$:
\begin{align*}
    (\tilde{\be} - \m)^\top\Sw^{-1}(\tilde{\be} - \m)
    =
    \kappa^2
    \frac{\bep^\top}{\Vert\bep\Vert}
    (\Sw^{\frac12})^\top
    \Sw^{-1}
    \Sw^{\frac12}
    \frac{\bep}{\Vert\bep\Vert}
    = \kappa^2 \,,
\end{align*}
such that $\tilde{\be}$ lies on the surface constituting a confidence ellipsoid of $\be$'s posterior.
We therefore refer to this strategy as \textit{Surface TS}.
In the numerical experiments, we simply set $\kappa=2$.

TuRBO provides for batched acquisition simply by performing Thompson sampling multiple times independently.
We adopt this strategy to our surface-TS method to produce batched samples with similarly minimal cost.

\section{UCB and EI are Hard}

Having established that TS is easily globally optimizable, we now prove that the same is not true of the more classical EI and UCB acquisition functions.
We do this by comparing resolution of these acquisition functions to the classical NP-complete \texttt{MAXCUT} problem.
Briefly, given a graph with nodes $\mathcal{G}$ and edges $E$, \texttt{MAXCUT} involves finding the partition of these nodes such that the number of edges connecting nodes from different partitions is maximal.

\begin{theorem}
    %For $\kappa>0$, the general class of problems formed by UCB optimization is NP-hard under a generic design matrix and $\ell$.
    Assume $P(\be) = N(\mathbf{0},\mathbf{I})$ and $\sigma^2=1$.
    Let $(\mathcal{G}, E)$ be a simple graph with nodes $\mathcal{G}$ and edges $E$, with $|\mathcal{G}|=D\geq2$ and $|E| = N \geq 1$.
    Create a design matrix $\X\in\mathbb{R}^{2N\times D}$ with rows given by:
    $\pm t (\e_i + \e_j)$ for each $(i,j)\in E$, where $\e_i$ denote the $i$th standard basis vector and $t=\frac{1}{100D^3}$.
    Choose $\ell = \frac1{10D}$.
    Then:
    \begin{enumerate}
        \item The input solving the variance-maximizing acquisition problem given by
        \begin{equation}
            \mathbf{c} = 
            \argmax{\x\in[-\ell,\ell]^D}
            \sigma^2(\x)
            %(\x^*)^\top \Sigma \x
        \end{equation}
        lies on a vertex, i.e. $\mathbf{c}\in\{-\ell,\ell\}^D$.
        \item Treating all nodes whose entries in $\mathbf{c}$ share a sign as belonging to common sides of the cut solves \texttt{MAXCUT} on this graph.
        \item Let $\y = \mathbf{0}$. Then EI (with $y^*=0$) and UCB (with $\kappa>0$) reduce to variance maximization. 
    \end{enumerate}
\end{theorem}
\begin{proof}
%The proof strategy involves choosing the design points so that the leading $D\times D$ block of the posterior covariance $\Sw$ is $(\I+a\La)^{-1}$ for a small constant $a>0$, where $\La=\mathbf{D}+\mathbf{A}$ is the signless Laplacian of the graph, $\mathbf{D}$ is the diagonal matrix giving node degrees and $\mathbf{A}$ is the adjacency matrix.
%Because both the box $[-\ell,\ell]^D$ and the design points within it are small, the Dinkelbach reformulation of variance maximization is convex along each coordinate, so the maximum lies on a vertex $\sol\in\{-\ell,\ell\}^D$.
%At a vertex, the Neumann series $(\I+a\La)^{-1}\approx\I-a\La$ shows that the variance is, up to a constant and a small remainder, decreasing in $\sol^\top\La\sol$, which is $4\ell^2$ times the number of edges joining nodes with the same sign in $\sol$.
%Maximizing the variance therefore minimizes the number of uncut edges, i.e. solves \texttt{MAXCUT}, and EI and UCB reduce to variance maximization when $\y=\mathbf{0}$.
See Appendix \ref{ap:maxcut}.
\end{proof}

Theorem 2 means that any algorithm which solves UCB or EI for all spherical-linear analyses with a design matrix $\mathbf{X}\in\mathbb{R}^{2N\times D}$ could also be used to solve \texttt{MAXCUT} on any graph with $N$ edges and $D$ nodes, which suggests that no such algorithm will be computationally efficient over all possible designs\footnote{Nevertheless, we cannot make a definitive statement of hardness until the $P$ vs $NP$ problem is solved (we leave that as future work).}.

%\subsection{Certification}
%
%Let's start by ignoring intercept with:
%$\w|\y\sim N(\mw, \Sw)$.
%
%Now, the mean at a given point is:
%\begin{equation}
%    \mu(\x) = 
%    \frac{2 \sum_{p=1}^P m_p x_p + m_{P+1}(\Vert\x\Vert^2-1)}
%    {\Vert\x\Vert^2+1}
%\end{equation}
%and the variance:
%\begin{equation}
%    \sigma(\x) = 
%    \frac{
%    \left[4 \sum_{i=1}^P\sum_{j=1}^P S_{i,j} x_i x_j \right]
%    + \left[2 \sum_{i=1}^P S_{P+1,i} x_i (\Vert\x\Vert^2-1)\right]
%    + \left[ S_{P+1,P+1} (\Vert\x\Vert^2-1)^2 \right]
%    }
%    {(\Vert\x\Vert^2+1)^2}
%\end{equation}
%
%Let's start with entropy search, i.e. can we just maximize $\sigma^2(\x)$ globally?
%The Dinkelbach iterates for that look like:
%\begin{align*}
%    & \min_{\x\in[-1,1]^P} \left[4 \sum_{i=1}^P\sum_{j=1}^P S_{i,j} x_i x_j \right]
%    + \left[4 \sum_{i=1}^P S_{P+1,i} x_i (\Vert\x\Vert^2-1)\right]
%    \\ & + 
%    \left[ S_{P+1,P+1} (\Vert\x\Vert^2-1)^2 \right]
%    -
%    q
%    (\Vert\x\Vert^2+1)^2
%\end{align*}
%We rewrite that in the following notation:
%\begin{align*}
%    4\x^\top\Sw\x + 4 (\Vert\x\Vert^2-1) \sw^\top\x + s(\Vert\x\Vert^2-1) - q(\Vert\x\Vert^2+1)
%    \,.
%\end{align*}
%where we have abused notation somewhat in redefining $\Sw$ and then put the omitted entries in $\sw$ and $s$.
%
%\subsection{Approach 1: Split off norm then dualize}
%This gives:
%\begin{align*}
%    & \max_{\x,z}
%    4\x^\top\Sw\x + 4 (\Vert\x\Vert^2-1) \sw^\top\x + s(\Vert\x\Vert^2-1) - q(\Vert\x\Vert^2+1)
%    \\ & s.t.\quad
%    \x\in[-\ell,\ell]^P; \quad z = x^\top\x
%\end{align*}
%The Lagrangian is

\section{Numerical Experiments}

In this section we compare the proposed Surface TS method with the standard multi-start LogEI approach proposed by \citet{doumont2026still} as well as some classical, GP-based BO competitors.
We present three main kinds of experiments. 
First is the classical result given in BO research, which compares the median Best observed function Value (BoV) to the number of black-box queries made, with each query made sequentially.
Second is this same thing but with batches of 50 queries at a time.
These are reasonable experiments when the black-box is very expensive such that its evaluations are all that matter.
But oftentimes the black-box is only of moderate expense, such that overhead incurred by BO cannot be completely neglected.
To study this, we also include a third kind of experiment, which compares median BoV to walltime, both actual and synthetic (where each black-box evaluation is assumed to take $\nu$ seconds), which allows us to compare the performance of the BO methods under varying assumptions for black-box execution time. 

See Appendix \ref{ap:design} for more on the experimental design.

%\todo{duplicated:
%In the BO literature, it is common to justify extraordinary acquisition overhead by reference to an even more expensive black-box.
%However, many users have black-boxes of ``intermediate" computational expense, such that additional computational overhead incurred by the BO algorithm can't be totally ignored. 
%In this subsection, we study the performance of the candidate BO methods using total walltime rather than evaluations as the budget.
%}

\subsection{Classical BO Evaluation}

Table \ref{tab:main} shows the median BoV at 1,000 evaluations.
On Rover, our surface-TS method with MAP estimation achieves the best value at the end of the experiment. 
On MOPTA08, Lasso-DNA, and SVM, it performs about similarly to LogEI while being much faster, though both are outperformed by GP-based methods.
Both LogEI and surface-TS with MAP fit perform quite well on Ant.
On Hopper, both Vanilla BO and TuRBO strongly outperform the spherical-linear methods, which have similar performance. 
On HalfCheetah, LogEI with Fixed estimation is the winner.
However, our MAP/surface-TS wins on both latent space optimization benchmarks.
LogEI with Fixed is also the best performing linear method on Lasso-hard, not too far behind Vanilla BO and TuRBO.
Overall, in terms of final BoV, MAP/surface-TS is able to take the overall lead on 3 of the 10 benchmarks, tying the record of Vanilla BO and beating all other methods.
See Figure \ref{fig:full_evals} of the Appendix for the complete trajectories on all problems with uncertainty bounds.
The middle-bottom panel of Figure \ref{fig:vary_wall} looks at things from a different angle, showing the $[0,1]$-normalized performance of each method averaged across all functions.
We see that MAP/surface-TS is beaten only by BOCK from this perspective and is much faster.

\begin{table*}[t]
\caption{Median BoV at 1,000 evaluations, lower is better.
Best method in bold, second-best underlined.
First four methods are Spherical-Linear models, next three are GP baselines.
}
\vspace{0.5em}
\label{tab:main}
\centering\small\setlength{\tabcolsep}{5pt}
\begin{tabular}{lrrrrr}
\toprule
Problem & Rover & MOPTA08 & Lasso-DNA & SVM & Ant \\
\textit{Dimension} & \emph{60} & \emph{124} & \emph{180} & \emph{388} & \emph{888} \\
\midrule
Linear (MAP) + LogEI & -1.76 & 237 & 0.289 & 0.079 & \underline{188} \\
Linear (MAP) + surface TS & \textbf{-3.28} & 238 & 0.289 & 0.083 & 222 \\
Linear (Fixed) + LogEI & -2.68 & 236 & 0.294 & 0.061 & 846 \\
Linear (Fixed) + surface TS & -2.85 & 242 & 0.296 & 0.061 & 913 \\
\cmidrule(lr){1-6}
Vanilla BO & -2.86 & \textbf{215} & \textbf{0.285} & \textbf{0.050} & 348 \\
BOCK & \underline{-3.26} & \underline{218} & \underline{0.286} & \underline{0.057} & \textbf{-1007} \\
TuRBO & -2.34 & 229 & 0.286 & 0.073 & 675 \\
\midrule[\heavyrulewidth]
 Problem & Hopper & HalfCheetah & Osimertinib & Fexofenadine & Lasso-hard \\
\emph{Dimension} & \emph{33} & \emph{102} & \emph{256} & \emph{256} & \emph{1000} \\
\midrule
Linear (MAP) + LogEI & -1026 & 28.3 & -0.590 & \underline{-0.737} & 4.17 \\
Linear (MAP) + surface TS & -1051 & 97.4 & \textbf{-0.773} & \textbf{-0.740} & 2.42 \\
Linear (Fixed) + LogEI & -1020 & \textbf{-842} & -0.581 & -0.597 & 1.75 \\
Linear (Fixed) + surface TS & -1029 & -31.1 & -0.618 & -0.665 & 3.05 \\
\cmidrule(lr){1-6}
Vanilla BO & \underline{-2002} & 221 & -0.629 & -0.649 & \underline{1.67} \\
BOCK & -1046 & \underline{-702} & \underline{-0.729} & -0.732 & 2.03 \\
TuRBO & \textbf{-2433} & 327 & -0.629 & -0.654 & \textbf{1.53} \\
\bottomrule
\end{tabular}

\end{table*}

\subsection{Walltime BO Evaluation}

We now discuss the results of the walltime analyses. 
The analysis with actually measured walltimes is presented in Figure \ref{fig:obs_wall} of the Appendix. 
We find there that on Rover and SVM the Fixed hyperparameter surface-TS method outperforms all others throughout the part of the space it was deployed.
On MOPTA08, Hopper, HalfCheetah and Lasso-hard it is dominant for smaller walltimes.
On the other hand, the MAP approach is better on Ant and the two latent-space problems.
TuRBO also performs well on Lasso-DNA, Lasso-hard and Hopper, but at higher walltimes than do the linear methods.
In particular, only on HalfCheetah does the multistart LogEI approach significantly outperform the surface-TS methods at any point, and even then its performance stalls and the best BoVs are achieved by our approach.

\begin{figure}
    \centering
    \includegraphics[width=0.99\linewidth]{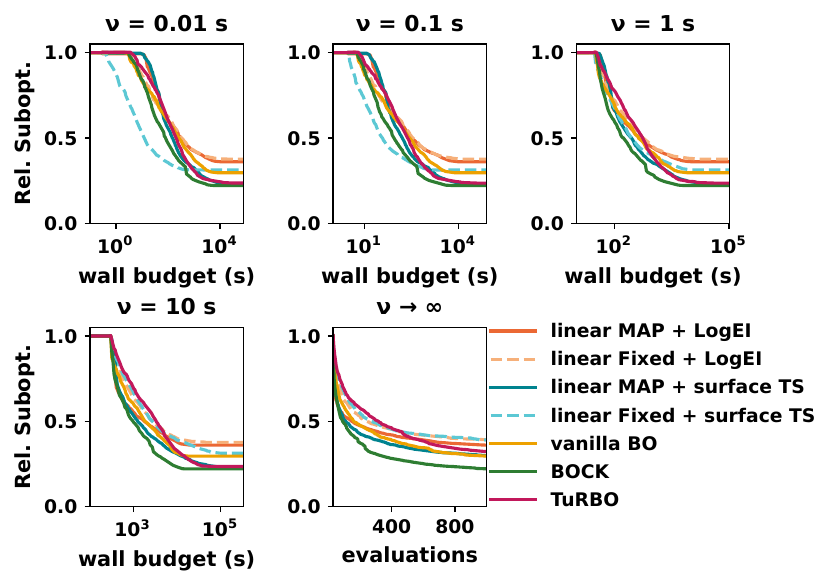}
    \caption{Average model performance across all problems with budget given by walltime; lower is better.}
    \label{fig:vary_wall}
\end{figure}

Next we discuss the synthetic walltime results, which are summarized in Figure \ref{fig:vary_wall}.
We see that when $\nu$ is small, the Fixed/surface-TS method has the best performance for budgets less than about $1{,}000$ seconds, while as the walltime increases, the MAP/surface-TS becomes worth the additional computational effort.
Figure \ref{fig:all_splot} of the Appendix shows the best-performing method as a function of $\nu$ and total walltime. 
On all problems aside from Ant, the Fixed/surface-TS method dominates the lower-left quadrant, when quick answers are needed from relatively cheap black-boxes.

\begin{figure}
    \centering
    \includegraphics[width=0.99\linewidth]{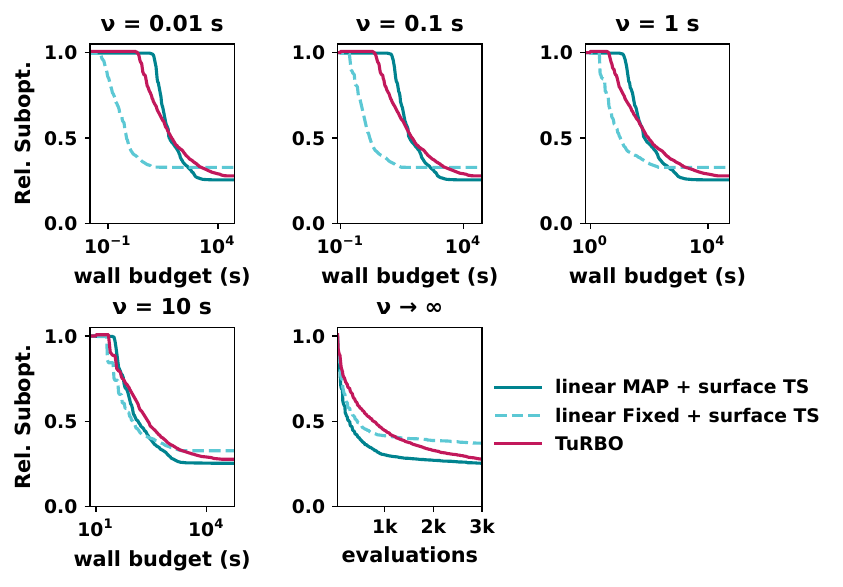}
    \caption{Average model performance across all problems with batch size of 50; lower is better.}
    \label{fig:batched}
\end{figure}

On the batch size 50 problems (Figure \ref{fig:batched}), we find that for $\nu<1\,\mathrm{s}$ the Fixed/surface-TS method performs best, while for $\nu>10\,\mathrm{s}$ the MAP/surface-TS performs best.

\section{Discussion}

\textbf{Summary:}
In this article, we surveyed the prospects for global optimization of acquisition functions associated with Spherical-Linear Bayesian Optimization.
We found that though multi-start local search can fail for Thompson sampling, an efficient globally optimal algorithm is available via Dinkelbach iteration.
Unfortunately, we also found that Thompson sampling is not very informative as the dimension of the problem grows, and proposed surface-TS to mitigate this.
We next demonstrated that global optimization of UCB and EI on SLBO can be used to solve \texttt{MAXCUT}, an NP-complete problem, and thus we do not expect an efficient general algorithm to exist for these acquisitions.

\textbf{Conclusions:}
Despite the apparent simplicity of using a linear model in a transformed space, actually carrying out the acquisition problem can be computationally difficult in this model class.

\textbf{Future Work:}
Though we have all but ruled out a general efficient algorithm for global UCB and EI optimization, this does not mean that these problems will often be intractable in practice.
It would be interesting to conduct a numerical study to investigate this empirical question.
Additionally, we are intrigued by the possibility of developing specialized local-search algorithms which are possibly more efficient than L-BFGS-B on this problem class.

\subsection*{AI use statement}
In this work, we used generative AI tools for implementation (code generation and editing), figure creation, to provide critical ingredients for proving mathematical claims, and for checking mistakes in the writeup.
We did not use generative AI tools to help develop theoretical models or conceptual frameworks, assist in the writing of proofs, propose or refine hypotheses, design or provide feedback on research methodology or experiments, assist with translation, clean and reformat datasets, support qualitative and thematic data analysis, draft any text in this article, or interpret results. 
The mathematical proofs presented in this work were all written manually and are based in part on output from various commercial LLMs including Claude Fable 5.1 and ChatGPT 6.1 Sol.
In all cases, the exact arguments presented here were synthesized by the authors after iteration with multiple LLM generated proofs and independent reflection.
We take full responsibility for the final content of this work.

%\subsection*{Reproducibility statement}
%Code implementing this method and sufficient to reproduce all results figures is provided in the supplementary material.
%Appendix B gives all details necessary for reproducing our experimental results.

\bibliographystyle{apalike}
\bibliography{main}

\clearpage
\appendix
\thispagestyle{empty}

% Supplementary material: To improve readability, you must use a single-column format for the supplementary material.
\onecolumn
\aistatstitle{
Acquisition for Spherical-Linear Bayesian Optimization on the Cube: \\
Supplementary Materials}

\section{Proofs}

\subsection{Proof of Theorem 1 (algebraic form of $P([-\ell,\ell]^D)$).}
\label{ap:algebra}

\textbf{Image Contained in Set $(\implies)$:}
Take any element $\x\in[-\ell,\ell]^D$.
Then 
\begin{equation}
    P(\x) = 
    \begin{bmatrix}
        \frac{2\x^\top}{\Vert\x\Vert^2+1}, 
        \frac{\Vert\x\Vert^2-1}{\Vert\x\Vert^2+1}
    \end{bmatrix}^\top
    :=
    \begin{bmatrix}
        \y^\top, s
    \end{bmatrix}^\top
    \,.
\end{equation}
That $\Vert(\y,s)\Vert=1$ is by construction and easily verified:
\begin{equation*}
    \Vert 
    P(\x)
    \Vert = 
    \frac
    {
    \sqrt{
    4
    \Vert
    \x
    \Vert^2
    +\Vert\x\Vert^4-2\Vert\x\Vert^2+1
    }}
    {\Vert\x\Vert^2+1}
    =
    \frac
    {
    \sqrt{
    (\Vert\x\Vert^2+1)^2
    }}
    {\Vert\x\Vert^2+1}
    = 1\,.
\end{equation*}
That $|y_i|\leq \ell(1-s)$ follows from noting first that since $s = \frac{\Vert\x\Vert^2-1}{\Vert\x\Vert^2+1}$, then $1-s=\frac{2}{\Vert\x\Vert^2+1}$. 
Therefore:
\begin{equation*}
    |y_i|=\left|\frac{2\x_i}{\Vert\x\Vert^2+1}\right|
    =
    |x_i|(1-s)\leq\ell(1-s)
    \,.
\end{equation*}

Finally $s\leq1-\epsilon$.
$s = \frac{\Vert\x\Vert^2-1}{\Vert\x\Vert^2+1}$ and $a\to\frac{a-1}{a+1}$ has derivative $\frac{2}{(1+a)^2}$ and is thus increasing, so $s$ increases in $\Vert\x\Vert^2$.
The maximum $\Vert\x\Vert^2 = D\ell^2$, so $s\leq \frac{D\ell^2-1}{D\ell^2+1}<1$ and can thus achieve $s\leq1-\epsilon$ for $\epsilon=1-\frac{D\ell^2-1}{D\ell^2+1}>0$.

\noindent\textbf{Set Contained in Image ($\impliedby$):}
We prove this by showing that the preimage of the purported expression under $P$ for $\Omega$ is contained in $[-\ell,\ell]^D$.

To this end, take any pair $(\y,s)$ such that $\Vert(\y,s)\Vert=1$, $|y_i|\leq \ell(1-s)$, and $s\leq 1-\epsilon$.
The inverse transformation gives $\x:=P^{-1}([\y^\top,s]^\top)=\frac{\y}{1-s}$, and the second condition of the previous sentence gives us that $|x_i|=\frac{|y_i|}{1-s}\leq \ell$ whenever $s<1$, which we have also assumed.
The unit-norm constraint is not needed for this direction.

\subsection{Proof of Corollary 1 (Geometric interpretation of $P([-\ell,\ell]^D)$).}
\label{ap:geometry}

We begin by demonstrating that 
$
\mathcal{H}
= 
\{(\y,s)\in\mathbb{R}^{D+1} : |y_d|\leq \ell(1-s), s\leq 1\}
-\n\,.
$
\begin{align*}
    & \mathcal{H} = 
    \cone(
    \{
    (\ell \bv, -1) : \bv\in\{-1,1\}^D
    \}
    )
    \\ & =
    \{
    t(\y,-1) : t\geq 0, \y\in\conv\left(\ell\{-1,1\}^D\right)
    \}
    \\ & =
    \{
    (\y,-t) : t\geq 0, \y\in t[-\ell,\ell]^D
    \}
    \\ & = 
    \{
    (\y,s-1) : s\leq 1, \y\in (1-s)[-\ell,\ell]^D
    \}
    \\ & = 
    \{
    (\y,s) : s\leq 1, \y\in (1-s)[-\ell,\ell]^D
    \}
    -\n
    \,,
\end{align*}
since $\conv(\ell\{-1,1\}^D)=[-\ell,\ell]^D$.

Then taking $(\mathcal{H}+\n)\cap S_D$ means we get the constraints from Theorem 1, namely:
\begin{equation}
    \{(\y,s) : |y_i|\leq \ell(1-s), s\leq 1, \Vert(\y,s)\Vert=1\}\,.
\end{equation}

\subsection{Proof of Theorem 2 (Solving \texttt{MAXCUT} via EI/UCB Optimization).}
\label{ap:maxcut}

This section establishes that an algorithm which can globally solve the UCB or EI subproblems associated with a spherical-linear kernel with any design matrix $\X\in\mathbb{R}^{2N\times D}$ and response $\y\in\mathbb{R}^{2N}$ can be used to solve \texttt{MAXCUT} on a graph with $N$ edges and $D$ nodes.
It proceeds by establishing this to be the case with the variance-maximizing acquisition, and then demonstrates how to export this to EI/UCB.

At a high level, the proof strategy operates by noting that the objective in variance maximization involves maximizing the ratio of a quadratic form and a quartic.
We engineer design points such that the quadratic form approximates the signless Laplacian matrix of the graph, and we choose $\ell$ to be very small such that the quartic in the denominator contributes little to the cost relative to the quadratic form.

For simplicity, this proof is presented without an intercept term.

\paragraph{Part 1: Optima are on the Vertices.} We begin by developing the predictive variance expression.
The posterior variance of the linear regression coefficients is given in general by:
\begin{equation}
    \Sw = (P(\X)^\top P(\X) + \I)^{-1}\,,
\end{equation}
and we have that
\begin{equation}
    P(\X)^\top P(\X) = \sum_{i=1}^{2N} P(\x_{i}) P(\x_i)^\top \,.
\end{equation}
We begin by investigating the contribution to this sum of an individual term, where we denote the sign by $s\in\{-1,1\}$:
\begin{align*}
    P(s t (\e_i+\e_j)) P(s t (\e_i+\e_j))^\top = 
    \frac1{(t^2\Vert\e_i+\e_j\Vert^2+1)^2}
    \begin{bmatrix}
        4 t^2(\e_i+\e_j)(\e_i+\e_j)^\top & s \mathbf{a} \\ 
        s\mathbf{a} & (t^2\Vert\e_i+\e_j\Vert^2-1)^2
    \end{bmatrix}
    \,.
\end{align*}
where $\mathbf{a} = 2 t (t^2\Vert\e_i+\e_j\Vert^2-1)(\e_i+\e_j)$.
Therefore, summing pairs with opposite signs yields:
\begin{equation}
    \frac2{(2t^2+1)^2}
    \begin{bmatrix}
        4t^2(\e_i+\e_j)(\e_i+\e_j)^\top & \mathbf{0} \\ 
        \mathbf{0}^\top & (2t^2-1)^2
    \end{bmatrix}\,,
\end{equation}
such that 
\begin{equation}
    P(\X)^\top P(\X) =
    \frac2{(2t^2+1)^2}
    \begin{bmatrix}
        4t^2\sum_{(i,j)\in E}
        (\e_i+\e_j)(\e_i+\e_j)^\top & \mathbf{0} \\ 
        \mathbf{0}^\top & N(2t^2-1)^2
    \end{bmatrix}
    :=
    \begin{bmatrix}
        a \La & \mathbf{0} \\
        \mathbf{0}^\top & b
    \end{bmatrix}
    \,,
\end{equation}
where we have defined 
$
a = \frac{8t^2}{(2t^2+1)^2}
$
and
$
b = \frac{2 N (2t^2-1)^2}{(2t^2+1)^2}
$
and
$
\La = \sum_{(i,j)\in E} (\e_i+\e_j)(\e_i+\e_j)^\top
$.
Then, 
\begin{equation}
    \Sw =
    \begin{bmatrix}
        (a\La+\I)^{-1} & \mathbf{0} \\ 
        \mathbf{0}^\top & \frac{1}{b+1}
    \end{bmatrix}
    \, \quad \textrm{and}\quad
    \sigma^2(\x) = P(\x)^\top\Sw P(\x)\,.
\end{equation}

\vspace{1em}

Now we investigate the optima of $\sigma^2(\x)$, starting by developing further its expression:
\begin{equation}
    \sigma^2(\x) = 
    \frac
    {
    4 \x^\top (a\La+\I)^{-1}\x +
     \frac{1}{b+1} (\Vert\x\Vert^2-1)^2
     }
    {(\Vert\x\Vert^2+1)^2}
    =
    \frac
    {
    4 \x^\top (a\La+\I)^{-1}\x}
    {(\Vert\x\Vert^2+1)^2}
    +
    \frac{
     \frac{1}{b+1} (\Vert\x\Vert^2-1)^2
    }
    {(\Vert\x\Vert^2+1)^2}
    \,.
\end{equation}
Noting that $(\Vert\x\Vert^2-1)^2=(\Vert\x\Vert^2+1)^2-4\Vert\x\Vert^2$, we obtain:
\begin{align}
    & \sigma^2(\x) = 
    \frac
    {
    4 \x^\top (a\La+\I)^{-1}\x}
    {(\Vert\x\Vert^2+1)^2}
    +
    \frac{
     \frac{1}{b+1} \left((\Vert\x\Vert^2+1)^2-4\Vert\x\Vert^2\right)
    }
    {(\Vert\x\Vert^2+1)^2}
    =
    \frac
    {
    4 \x^\top (a\La+\I)^{-1}\x}
    {(\Vert\x\Vert^2+1)^2}
    -
    \frac{
      \frac{4}{(b+1)}\Vert\x\Vert^2
    }
    {(\Vert\x\Vert^2+1)^2}
    +\frac{1}{b+1}
    \\ & = 
    4 
    \frac
    {
    \x^\top (a\La+\I)^{-1}\x
  -\frac1{(b+1)}\Vert\x\Vert^2
    }
    {(\Vert\x\Vert^2+1)^2}
    +\frac{1}{b+1}
    =
    4\frac{\x^\top \left(
    (a\La+\I)^{-1}-\frac1{b+1}\I
    \right)
    \x
    }
    {(\Vert\x\Vert^2+1)^2}
    +\frac{1}{b+1}
    \,.
\end{align}
Therefore maximizing $\sigma^2(\x)$ is equivalent to maximizing
\begin{align}
    \tilde{\sigma}^2(\x) = 
    \frac{\x^\top \Q
    \x
    }
    {(\Vert\x\Vert^2+1)^2}\,,
    \quad\textrm{where}\quad
    \Q
    =
    (a\La+\I)^{-1}-\frac1{b+1}\I
    \,.
\end{align}

The numerator encourages $\x$ to be large, while the denominator wants it to be small.
It will therefore be useful to bound the relative influence of each, and to this end we consider the norm of $\La$. 
Since
\begin{equation}
    \x^\top\La\x = 
    \sum_{(i,j)\in E} (x_i+x_j)^2 
    \leq  2\sum_{(i,j)\in E} (x_i^2 + x_j^2 )
    \leq 2 \sum_{i < j} (x_i^2+x_j^2)
    = 2(D-1)\Vert\x\Vert^2 \,,
\end{equation}
we have that $\mathbf{0}\preceq \La \preceq 2(D-1)\I$ where $\preceq$ gives the ordering induced by the positive semidefinite cone on $\mathbb{R}^{D\times D}$.

Now choose $t=\frac{1}{100D^3}$.
Then $a = \frac{8t^2}{(2t^2+1)^2}<8t^2 = \frac{8}{10000 D^6} < \frac{1}{2D^3}$.
Therefore, $\Vert a\La\Vert \leq \frac{2(D-1)}{2D^3}=\frac{D-1}{D^3}$.
For $D\geq 2$ this is maximized at $D=2$ at which point it evaluates to $\frac{1}{8}$.
This implies that $\Vert\I + a\La\Vert \leq \frac{9}{8}$, and together with $a\La$'s positive semi-definiteness we obtain
$\frac{8}{9}\I \preceq (\I+a\La)^{-1} \preceq \I$.

Now we bound the final cell, $\frac{1}{b+1}$.
Since $\tau\to\frac{(2\tau^2-1)^2}{(2\tau^2+1)^2}$ is unimodal and achieves its minimum at $\tau=\frac1{\sqrt{2}}$, we have that it is decreasing on the interval $[0,\frac1{10}]$.
At $\tau=\frac{1}{10}$, it gives value $\frac{2401}{2601}>\frac{1}{2}$.
With our choice of $t$, we therefore have that 
$
b = \frac{2 N (2t^2-1)^2}{(2t^2+1)^2} > 2 N \frac{1}{2} \geq 1
$
such that
$
\frac{1}{b+1}< \frac12.
$

We now deploy these two facts to bound 
$(a\La+\I)^{-1}-\frac1{b+1}\I$.
Given the just-discussed bound on $b$, we have
$(a\La+\I)^{-1}-\frac1{b+1}\I
\succ
(a\La+\I)^{-1}-\frac12\I
\succeq \frac{7}{18}\I
$,
which together with the positive-definiteness of $\frac1{b+1}\I$ 
yields
$
\frac{7}{18}\I \preceq 
%(a\La+\I)^{-1}-\frac1{b+1}\I
\Q
\preceq 
\I
$.

We can convert this into a bound on the optimal value of $\tilde{\sigma}^2(\x)$. 
Since $\Vert\mathbf{Q}\Vert\leq 1$, 
\begin{equation}
    \tilde{\sigma}^2(\x) =
    \frac{\x^\top\mathbf{Q}\x}{(\Vert\x\Vert^2+1)^2}
    \leq 
    \frac{\Vert\x\Vert^2}{(\Vert\x\Vert^2+1)^2}
    \leq
    \Vert\x\Vert^2
    \,,\quad\textrm{ and since }
    \x\in[-\ell,\ell]^D,\quad
    \Vert\x\Vert^2\leq D\ell^2 
    \,.
\end{equation}
Therefore, $\max_{\x\in[-\ell,\ell]^D} \tilde{\sigma}^2(\x) \leq D\ell^2=\frac{D}{100D^2}\leq\frac{1}{200}$.

This is useful because of the Dinkelbach equivalence:
\begin{equation}
    \argmax{\x\in[-\ell,\ell]^D} \tilde{\sigma}^2(\x)
    =
    \argmax{\x\in[-\ell,\ell]^D}
    \x^\top\Q\x - q^*
    (1+\Vert\x\Vert^2)^2
    \,,
\end{equation}
where $q^* = \max_{\x\in[-\ell,\ell]^D}\tilde{\sigma}^2(\x)$.
We now seek to show that this Dinkelbach problem is convex marginally along each coordinate, and thus that maximization over it goes to the boundary.
To this end, we develop its second partial derivatives:
\begin{align*}
    & \frac{\partial^2}{\partial x_p^2}
    \left(
    \x^\top\Q\x - q^*
    (1+\Vert\x\Vert^2)^2
    \right)
    =
    2 Q_{pp}
    - q^*
    \left(
    4(1+\Vert\x\Vert^2)+8x_p^2
    \right)
    \\ & \geq 
    2\frac{7}{18} - \frac{1}{200}\left(4\left(1+\frac1{200}\right)+\frac{8}{20}\right)
    = \frac{7}{9} - \frac{221}{10000} > 0
    \,.
\end{align*}

Therefore, for all other $\x$ held fixed, the coordinate subproblem with respect to $x_p$ must have solution on a boundary.
It must therefore be the case that all coordinates of the optimum are at the boundary, lest we encounter a contradiction.

\paragraph{Part 2: Vertex Optimization is Equivalent to \texttt{MAXCUT}.}

Let us now consider a candidate vertex-valued solution $\sol\in\{-\ell,\ell\}^D$, and evaluate the cost there:
\begin{equation}
    \tilde{\sigma}^2(\sol) =
    \frac{\sol^\top\Q\sol}{(\Vert\sol\Vert^2+1)^2}
    =
    \frac{1}
    {(D\ell^2+1)^2}
    \sol^\top\Q\sol
    \,.
    %\propto
    %\frac{1}
    %{(D\ell^2+1)^2}
    %\sum_{(i,j)\in E} \text{\todo{complete this expression}}
\end{equation}

Recall that $\Q = (a\La + \I)^{-1}-\frac{1}{b+1}\I$; the second term is constant over vertices, so we drop it in what follows. 
Meanwhile, the \texttt{MAXCUT} objective for the trivial weighted graph with all weights given by $a$ has quadratic form given by $\I - a\La$, and \texttt{MAXCUT} on this graph is equivalent to that on the concomitant simple graph.
This motivates us to consider the Neumann series
\begin{equation}
    (\I+a\La)^{-1} = \sum_{t=0}^\infty (-a)^t \La^t
    = \I - a\La + \sum_{t=2}^\infty (-a)^t \La^t
    := \I - a\La + \mathbf{R}\,,
\end{equation}
where $\mathbf{R}$ is a remainder term.
We now develop the norm of this remainder:
\begin{align*}
    \Vert\mathbf{R}\Vert = 
    \left\Vert\sum_{k=2}^\infty (-a)^k \La^k\right\Vert
    \leq
    \left\Vert a^2\La^2 \sum_{k=0}^\infty a^k \La^k\right\Vert
    \leq
    a^2\Vert \La \Vert^2
    \sum_{k=0}^\infty a^k \left \Vert \La^k\right\Vert
    =
    \frac{a^2\Vert\La\Vert^2}{1-a\Vert\La\Vert}\,.
\end{align*}
Since, from above, $\Vert\La\Vert\leq 2(D-1) < 2D$, and $a\Vert\La\Vert\leq \frac18$, and furthermore, since $a = \frac{8t^2}{(2t^2+1)^2}< 8t^2 = \frac{1}{1250 D^6} < \frac1{8D^3}$, we have:
\begin{align*}
    \Vert\mathbf{R}\Vert< \frac{a^2 4D^2}{1-\frac18}
    = \frac{32}{7} D^2 a^2
    < \frac{32}{7} D^2\frac{1}{8D^3} a
    = \frac{4}{7} \frac{a}{D}
    < \frac{a}{D} \,.
\end{align*}

Therefore, $\mathbf{c}^\top\mathbf{R}\mathbf{c} \leq \Vert \mathbf{c}\Vert^2 \Vert\mathbf{R}\Vert \leq \ell^2 a$, such that for any two candidate solutions $\mathbf{c}_1,\mathbf{c}_2$, $|\mathbf{c}_1^\top\mathbf{R}\mathbf{c}_1 - \mathbf{c}_2^\top\mathbf{R}\mathbf{c}_2|\leq 2\ell^2a$.

On the other hand, 
$
\mathbf{c}^\top \La\mathbf{c} = \sum_{(i,j)\in E} (c_i+c_j)^2
$.
If $c_i=c_j$, which we interpret as nodes $i$ and $j$ belonging to the same side of the cut, then $(c_i+c_j)^2=4\ell^2$. 
If not, which means they belong to different sides of the cut, then $(c_i+c_j)^2 = 0$, such that $\frac{\mathbf{c}^\top\La\mathbf{c}}{4\ell^2}$ is measuring the number of \textit{uncut} edges, which we subtract from the total edge number $N$ to obtain the number of cut edges.
We thus have that 
$
\mathbf{c}^\top \La\mathbf{c} = 4N\ell^2 - 4\ell^2\textrm{cut}(\mathbf{c})
$, so:
\begin{align*}
    \mathbf{c}^\top \Q\mathbf{c} = 
    \mathbf{c}^\top(\I - a\La + \mathbf{R})\mathbf{c} = 
    \Vert\mathbf{c}\Vert^2
    - a(4N\ell^2 - 4\ell^2\textrm{cut}(\mathbf{c}))
    + \mathbf{c}^\top\mathbf{R}\mathbf{c}
    = \textrm{const} + 4a\ell^2\textrm{cut}(\mathbf{c}) + \mathbf{c}^\top\mathbf{R}\mathbf{c}
    \,.
\end{align*}
Now let $\sol_*\in\argmax{\sol\in\{-\ell,\ell\}^D}\sol^\top\Q\sol$ and $\sol_1$ be a suboptimal solution.
Then:
\begin{align*}
    \mathbf{c}_*^\top \Q\mathbf{c}_*
    >
    \mathbf{c}_1^\top \Q\mathbf{c}_1
    \iff
    4a\ell^2\cut(\sol_*) + \sol_*^\top\mathbf{R}\sol_* > 4a\ell^2\cut(\sol_1)  + \sol_1^\top\mathbf{R}\sol_1
    \iff 
    \cut(\sol_*) > \cut(\sol_1) + 
    \frac{\sol_1^\top\mathbf{R}\sol_1 - \sol_*^\top\mathbf{R}\sol_*}{4a\ell^2}
    \,.
\end{align*}
but we have that 
$
    \left|\frac{\sol_1^\top\mathbf{R}\sol_1 - \sol_*^\top\mathbf{R}\sol_*}{4a\ell^2}\right|
    \leq
    \frac12
$, which gives us that 
$
\cut(\sol_*) - \cut(\sol_1) >  -\frac12\,.
$
But $\cut(\sol_*) - \cut(\sol_1)\in\mathbb{Z}$, so in fact
$\cut(\sol_*) - \cut(\sol_1) \geq  0 \iff \cut(\sol_*) \geq \cut(\sol_1)$.
Therefore, $\sol_*\in\argmax{\sol} \cut(\sol)$.

\paragraph{Part 3: Extension to UCB/EI.}

We now assume that $\y =\mathbf{0}$. 
Then, the posterior mean $\mu(\x) = P(\mathbf{x})^\top \Sw P(\X)^\top \y = 0$.

Start with the UCB objective. 
For $\kappa>0$, we then have UCB given by $\mu(\x) + \kappa \sigma(\x) = \kappa\sigma(\x)$.
Maximizing this is equivalent to maximizing $\sigma^2(\x)$, which we have just demonstrated may be used to solve \texttt{MAXCUT}.

EI is given by 
$
(\mu(\x) - y^*) \Phi(\frac{\mu(\x) - y^*}{\sigma(\x)}) + 
\sigma(\x) \phi(\frac{\mu(\x) - y^*}{\sigma(\x)})\,.
$
Take $y^* = 0$. 
We have therefore that EI is given by
$ 0 \Phi(0) +  \sigma(\x) \phi(0) \propto \sigma(\x)\,, $
which again results in variance maximization.

\section{Additional Experimental Details}

\subsection{Experimental Design}
\label{ap:design}

We now describe in turn the BO methods we consider, the benchmarks on which they are deployed, and the strategy for estimating their performance.

\subsubsection{Competitors}

We use ``Vanilla BO'' as implemented by \citet{hvarfner2024vanilla}, TuRBO \citep{eriksson2019scalable}, and BOCK \citep{oh2018bock} as classical GP baselines. 
As this paper is focused on spherical-linear BO, these are not meant to be the main comparison, but are provided for context.

We deploy SLBO based on the code provided by the authors \citep{doumont2026still}, but with hyperparameters fit two different ways.
First, following the original article, we perform MAP inference to learn scale, noise, and dimension-specific lengthscale parameters; this is denoted by \textit{linear MAP}.
Second, we port their later approach \citep{fan2026search}, which keeps lengthscales fixed and uses a rapid iteration to fit the remaining two parameters and which was developed in the context of spherical search regions, to the box; we refer to this method as \textit{linear Fixed}.

We compare our surface-TS acquisition strategy to that used originally in \citet{doumont2026still}, which simply uses BoTorch's default approach of initializing 4 multi-start L-BFGS-B searches at the most promising of 512 Sobol points together with 512 points near the optimum.
Our method for acquisition is denoted ``surface-TS'' and the default strategy ``LogEI'' in the graphics.

\subsubsection{Benchmarks}

We started with five benchmarks studied by \citet{doumont2026still}, and we used performance on these problems in developing our method. 
Namely, these were the 60-dimensional Rover trajectory planning problem originally introduced by \cite{wang2018batched}, the 124-dimensional MOPTA08 vehicle design problem \citep{jones2008mopta}, the 180-dimensional Lasso-DNA problem \citep{vsehic2022lassobench}, the 388-dimensional SVM hyperparameter problem originally studied in \citet{eriksson2021high}, and the 888-dimensional MuJoCo Ant linear policy studied by \citet{mania2018simple}.
After primary development, we also added five additional benchmarks to validate generalizability.
Namely, we use two additional MuJoCo linear policy problems also from \citet{mania2018simple} (the 33-dimensional Hopper and the 102-dimensional HalfCheetah), the Osimertinib and Fexofenadine latent-space optimization tasks from \citet{fan2026search}, both 256-dimensional, and the LassoBench's synthetic ``hard'' problem in dimension 1,000 \citep{vsehic2022lassobench}.

\subsubsection{Performance Estimation}

Each run is initialized with 30 Sobol points, which are shared by all competitors for a given seed. 
In the primary set of experiments, we ran all methods for 970 sequential evaluations. 
Subsequently, we extended the length of experiments with faster methods so as to match the walltime of the most expensive method.
This entire procedure is repeated 20 times for different seeds.
We had access to GTX 1080 Tis and 8-core Ice Lake CPUs; each method was timed on each and assigned to the fastest.

For walltime evaluation, we considered two metrics.
First was the actual execution time measured as we executed the study. 
This has the benefit of being the actual quantity experienced by an analyst, but suffers from interference by external factors, such as node concurrency in our shared computing environment.
We therefore also developed a synthetic walltime calculation as follows.
Actual walltime for each iteration may be expressed as a sum of surrogate fitting time $t^{\textrm{fit}}$, acquisition time $t^{\textrm{acq}}$, black-box evaluation time $t^{f}$, and a negligible overhead. 
We record these first three quantities separately.
Then we replace the actually observed $t^f$ by a common term $\nu$, with $\nu$ varied from 0 to 1,000 seconds.
This allows us to simulate the execution time we would have observed had the cost of evaluating the black-box actually been $\nu$ seconds for any $\nu$ while running only a single experiment.
For batch acquisition, we consider both perfect parallelism, such that a round of 50 evaluations still costs only $\nu$ seconds, as well as fully sequential evaluation, such that it costs $50\nu$ seconds.

\subsection{Additional Figures}

\begin{figure}
    \centering
    \includegraphics[width=0.95\linewidth]{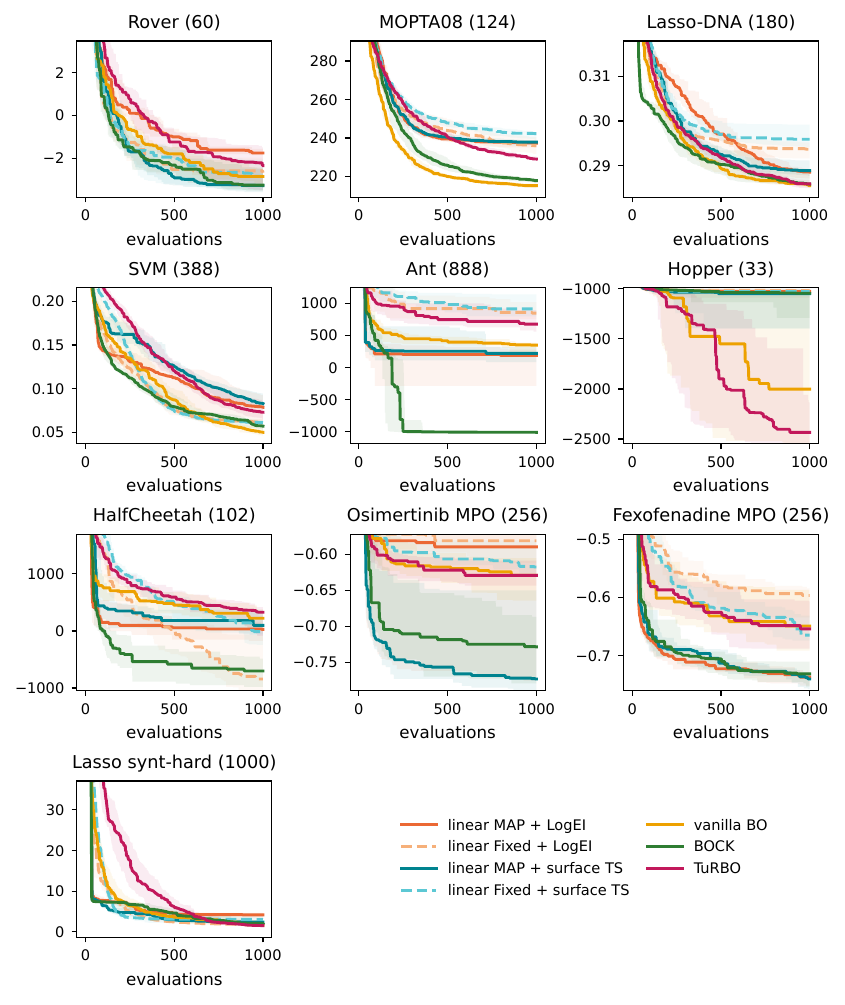}
    \caption{Full results with budget given by evaluations. Shaded regions are interquartile range.}
    \label{fig:full_evals}
\end{figure}

\begin{figure}
    \centering
    \includegraphics[width=0.95\linewidth]{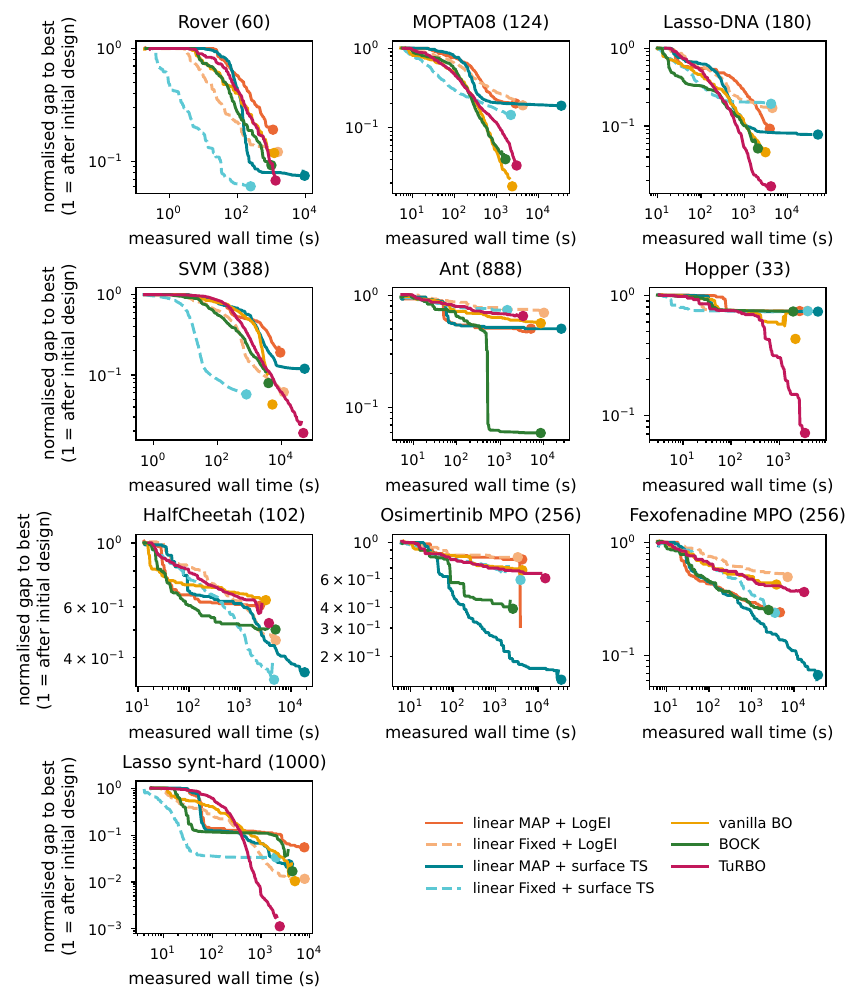}
    \caption{Full results of empirically observed walltime versus BoV.}
    \label{fig:obs_wall}
\end{figure}

\begin{figure}
    \centering
    \includegraphics[width=0.95\linewidth]{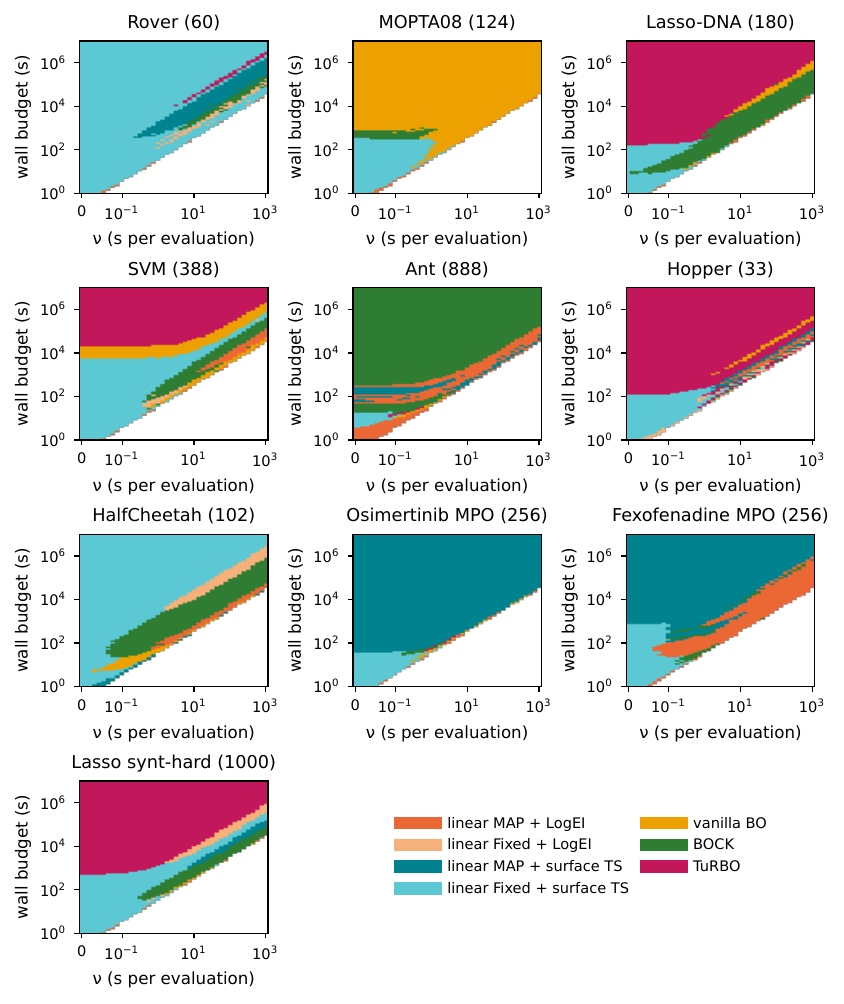}
    \caption{Full results showing the best performing method as a function of $\nu$, the time it takes a single black-box call to resolve, and the total walltime budget.
    The shaded color indicates the method of choice for a given combination.
    }
    \label{fig:all_splot}
\end{figure}

Figure \ref{fig:full_evals} gives the complete trajectories for all test problems when using black-box evaluations as the limiting budget.
Figure \ref{fig:obs_wall} on the other hand gives these same trajectories but where the budget is given by the observed walltime as actually measured on the compute nodes of our HPC system.
Since this is shared with other users, it is possible that there is interference from concurrent computation.
Figure \ref{fig:all_splot} shows the best-performing method for a given combination of budgets and black-box execution time.

\end{document}